\documentclass[10pt,twocolumn,letterpaper]{article}

\usepackage{cvpr}
\usepackage{times}
\usepackage{epsfig}
\usepackage{graphicx}
\usepackage{amsmath}
\usepackage{amssymb}
\usepackage{subfigure}
\usepackage{verbatim}

\usepackage[pagebackref=true,breaklinks=true,letterpaper=true,colorlinks,
bookmarks=false]{hyperref}

\cvprfinalcopy 

\def\cvprPaperID{1608} 

\ifcvprfinal\fi
\begin{document}

\title{Mixture of Parts Revisited: Expressive Part Interactions for Pose
Estimation}

\author{Anoop Katti\\
Indian Institute of Technology Madras\\
Chennai, India- 600036\\
{\tt\small akatti@cse.iitm.ac.in}
\and
Anurag Mittal\\
Indian Institute of Technology Madras\\
Chennai, India- 600036\\
{\tt\small amittal@iitm.ac.in}
}

\maketitle

\begin{abstract}
Part-based models with restrictive tree-structured interactions for the Human
Pose Estimation problem, leaves many part interactions unhandled.
Two of the most common and strong manifestations of such unhandled interactions
are self-occlusion among the parts and the confusion in the localization of the
non-adjacent symmetric parts. 
By handling the self-occlusion in
a data efficient manner, we improve the performance of the basic Mixture of
Parts model by a large margin, especially on uncommon poses. Through addressing
the confusion in the symmetric limb localization using a combination of two
complementing trees, we improve the performance on all the parts by atmost
doubling the running time. Finally, we show that the combination of the two
solutions improves the results. We report results that are equivalent
to the state-of-the-art on two standard datasets. Because of maintaining the
tree-structured interactions and only part-level modeling of the base Mixture
of Parts model, this is achieved in time that is much less than the best
performing part-based model.
\end{abstract}

\section{Introduction}
Human Pose Estimation in a 2D image is the task of detecting the presence of
humans in the image and localizing their body parts. This problem is motivated
by its potentially enormous applicability in high-level vision tasks such as
Action Detection, Human Computer Interaction, Gesture Recognition,
automatic analysis of videos of people etc. 

A challenge unique to the human pose estimation problem is the large
articulation that characterizes the human body. The most successful approaches
are based on part-based models~\cite{Fischler:1973:RMP:1309264.1309318,
felzenszwalb2005pictorial, felzenszwalb2010object}. Here, the human body
is modeled as an articulation of deformable body parts, flexibly connected to
each other via spring-like connections. The appearance of each part is
modeled independently. Due to dividing the entire body as an articulation of
smaller parts, part-based model can handle a combinatorially large number of
articulations.

\begin{figure}[t]
\begin{center}
  \subfigure[]{\label{fig:soFMoP}\includegraphics[
  width=0.25\linewidth]{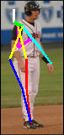}} 
  \subfigure[]{\label{fig:soOurs}
\includegraphics[width=0.25\linewidth]{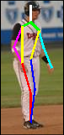}} \\
  \subfigure[]{\label{fig:loFMoP}\includegraphics[
  width=0.25\linewidth]{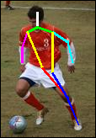}} 
  \subfigure[]{\label{fig:loOurs}
  \includegraphics[width=0.25\linewidth]{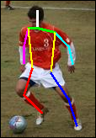}}
\end{center}
   \caption{Example pose estimation by Yang
\etal~\cite{yang2011articulated}(a, c) and our model (b, d). Erroneous pose
estimation in (a) is due to the unhandled self-occlusion (here, of the
upper-arm by the torso) and in (c) is due to the absence of a kinematic
constraint between the left and the right lower limbs.}
\label{fig:long}
\label{fig:motivation}
\end{figure}

Estimating pose while modeling all the interactions among the parts amounts to
inferencing on a loopy-graph, which is
intractable~\cite{koller2009probabilistic}. Yet, algorithms performing
approximate inference have been
proposed~\cite{gupta2008constraint, sigal2006measure, tian2010fast,
tran2010improved}, but at the cost of a high running time.
Therefore, in order to perform efficient pose estimation using Dynamic
Programming, the interactions between the parts are typically restricted to a
tree structure~\cite{felzenszwalb2005pictorial,
Fischler:1973:RMP:1309264.1309318, yang2011articulated} (Fig~\ref{fig:UCT}).
However,
the downside of such a restriction is that it fails to capture the interactions
between the non-adjacent parts. Two strong and
frequently occurring manifestations of such unhandled interactions are (i)
self-occlusion among the body parts producing large variations in
part appearances (\eg a fully visible upper arm vs an upper arm occluded by
the torso in the side-view, Fig~\ref{fig:soFMoP}) and
(ii) a confusion in the localization of the non-adjacent symmetric limbs due
to the absence of a kinematic constraint between them (\eg in the tree shown in
Fig~\ref{fig:UCT}, the left and the right legs are not kinematically
constrained. Therefore, predictions of both the legs may localize on one of the
legs as shown in Fig~\ref{fig:loFMoP}).

In order to perform inference tractably while still capturing higher order
interactions among parts, more recent works model
combinations of physically close body parts, instead of a single part.
For example, ~\cite{tian2012exploring,wang2013beyond} use latent nodes for the
combined parts to capture the higher order spatial relations. Each type
of any latent node prefers certain locations and types of its smaller
constituent observed nodes.
~\cite{gkioxari2013articulated, pishchulin2013poselet, pishchulin2013strong,
wang2011learning} use many larger rigid templates spanning
multiple parts called Poselets. Poselets handle part interactions by
capturing the variations in the appearance caused by them. Depending on which of
the poselets are activated, images are further processed to localize the parts.
Other methods to handle the higher order interactions perform
clustering in the pose space at semi-global~\cite{sapp2013modec} or
global~\cite{Johnson10,johnson2011learning} level and learn
cluster-specific deformable models.

The problem with all the above methods is that the variations in such
``superparts'' increase combinatorially as a function of the number of
constituent parts.
Therefore, when compared to part-level modeling, many more
latent-node-types/templates/clusters are required to handle such large
variations, consequently demanding a larger training data and a higher test
time.

Mixture of Parts (MoP) model, proposed by Yang and
Ramanan~\cite{yang2011articulated}, has been shown to be versatile
through its successful application in various problems of
computer vision~\cite{desai2012detecting, ramananparsing, yang2011articulated,
zhu2012face}. This is mainly due to the flexibility of modeling the
multi-modal appearances of parts by learning a mixture of templates per part,
instead of a single template. Particularly, Desai and
Ramanan's~\cite{desai2012detecting} method of
handling the self-occlusion within the MoP framework, while maintaining the
tree structured interactions with part-level modeling, is promising.

In our paper, we propose two modular and efficient improvements over Desai and
Ramanan~\cite{desai2012detecting} to handle (i) the self-occlusion in a more
data efficient manner and (ii) the confusion in localization of the
non-adjacent symmetric parts. The improvements are modular because each solution
can be applied independently, giving different strengths to the base MoP
model or in combination, giving the best performance; they are efficient because
they maintain the tree structured interactions and part-level modeling.

Specifically, for self-occlusion, we learn a mixture of templates
that capture the rotation normalized self-occlusion patterns in contrast to the
unnomrmalized patterns captured by Desai and Ramanan, thus increasing their data
efficiency. This gives especially large gains on uncommon poses like sports,
gymnastics and dance 

For addressing the confusion in the localization of non-adjacent symmetric
parts, we first propose a new tree (Fig~\ref{fig:LCT}) that localizes the
symmetric lower limbs with much less ambiguity and in some special scenarios,
even localizes the upper body parts better. Further, a generally applicable
solution that performs well on all the parts using a combination of the
traditionally used tree (Fig~\ref{fig:UCT}) and the new tree is also proposed.
These solutions provide a boost in the performance by atmost doubling the
running time.

We also show that the combination of the two solutions improves the results over
employing any one. 
On two standard datasets,
namely the LSP~\cite{Johnson10} and the IP~\cite{ramanan2006learning}, we
report results that are on par with the state-of-the-art methods in time that is
much faster than the best performing part-based
model~\cite{pishchulin2013strong}.

\section{Part-Based Model}
The part-based model has been shown to be very powerful since it can
handle large arbitrary articulations compared to a full-object
model~\cite{felzenszwalb2010object, felzenszwalb2005pictorial} . One popular
implementation of the part-based model is the Mixture of Parts
model~\cite{yang2011articulated}. In this model, the appearance
of every part is modeled using a mixture of templates, rather than a single
template. This makes the method more robust to variations in the appearance of
the parts. In this paper, we use the Mixture of Parts model as our baseline upon
which we build our ideas. 

In this section, we first review the Mixture of Parts
model. This is followed by a review of the approximations made by the
part-based model for efficient estimation of the pose and two issues arising
from it, namely the self-occlusions and the confusion in the localization of the
symmetric parts due to insufficient constraints. We also review Desai and
Ramanan's Phraselets~\cite{desai2012detecting} approach for deriving the
part mixtures to address the issue of self-occlusion.

\subsection{Mixture of Parts (MoP) Model}
Let $I$ be the given image and let $G = (V,E)$ be the MoP model,
where $V$ is the set of parts and $E$ is the set of pair-wise constraints
between the connected parts. Each part $i$ is parameterized by $(p_i,t_i)$,
where $p_i=(x_i,y_i)$ is the pixel location and $t_i$ is the mixture
type in the mixture of templates for part $i$. Let $(\mathbf{p},\mathbf{t})$
represent a pose configuration,
where $\mathbf{p} = [p_1 \ldots p_{|V|}]^T$ and $\mathbf{t} = [t_1 \ldots
t_{|V|}]^T$. Then the MoP model, parameterized by $(\mathbf{w},\mathbf{b})$,
scores a pose configuration $(\mathbf{p},\mathbf{t})$ on an image $I$ as:

\begin{eqnarray}
\label{eqn:MoP}
S(I,\mathbf{p},\mathbf{t}; \mathbf{w},\mathbf{b}) = 
\displaystyle \sum \limits_{i \in V} \left[ w_i^{t_i} \cdot \phi(I,p_i) \right]
+ \nonumber \\
\displaystyle \sum \limits_{(i,j) \in E} \left[ w_{i,j}^{t_i} \cdot
\psi(p_i-p_j) + b_{i,j}^{t_i,t_j} \right]
\end{eqnarray}

The first term in (\ref{eqn:MoP}) scores the matching of part-type specific
template $w_i^{t_i}$ to the HOG~\cite{dalal2005histograms} features
$\phi(I,p_i)$ extracted from the image $I$ at $p_i$.

The second term in (\ref{eqn:MoP}) enforces (a) part-type specific kinematic
constraints: $\psi(p_i-p_j) = [-dx, -dx^2, -dy, -dy^2]$, where $dx =
(x_i-x_j-\mu x_{i,j}^{t_i})$, $\mu x_{i,j}^{t_i}$ is the average
difference in the $x-$values between part $i$ and its parent $j$ in the training
images with type
$t_i$ for part $i$ and $dy$ is similarly defined; (b) type compatibility
constraints between a part and its parent: $b_{ij}^{t_it_j}$ scores the
compatibility of type $t_i$ of part $i$ and type $t_j$ of its parent $j$.

Given a test image $I$, the objective is to infer the max-scoring pose
configuration, i.e. $(\mathbf{p}^*,\mathbf{t}^*) = argmax \
S(I,\mathbf{p},\mathbf{t};\mathbf{w},\mathbf{b})$ or the set of pose
configurations above some threshold in the case of multiple persons in an image.
\\

\noindent \textbf{Learning:}
The model parameters, $(\mathbf{w},\mathbf{b})$, are learned using structured
SVM. Let $(\mathbf{p}^n,\mathbf{t}^n)$ be the ground-truth pose configuration of
the $n^{th}$ positive training image $I^n$. Then, $\beta =
(\mathbf{w},\mathbf{b})$ is obtained by solving:
\begin{eqnarray}
\label{eqn:SSVM}
 \displaystyle \arg \min_{\beta, \xi \geq 0} \frac{1}{2}\|\beta\|^2 + C
\displaystyle \sum_n \xi_n \\ \nonumber
s.t. \ \ \forall n \in pos \ \ S(I^n,\mathbf{p}^n,\mathbf{t}^n; \beta) \geq
1-\xi_n \\ \nonumber
\forall n \in neg,\ \forall p,t \ \ S(I^n,\mathbf{p},\mathbf{t}; \beta) \leq
-1+\xi_n
\end{eqnarray}
The above quadratic program solves for the lowest norm $\beta$ that scores
the ground-truth pose configurations in positive images above 1 and negative
images below -1. This is solved using the dual coordinate descent solver
of \cite{yang2011articulated}.

\subsection{Part Interactions}
\begin{figure}[]
\begin{center}
  \subfigure[]{\label{fig:UCT}\includegraphics[
  width=0.4\linewidth]{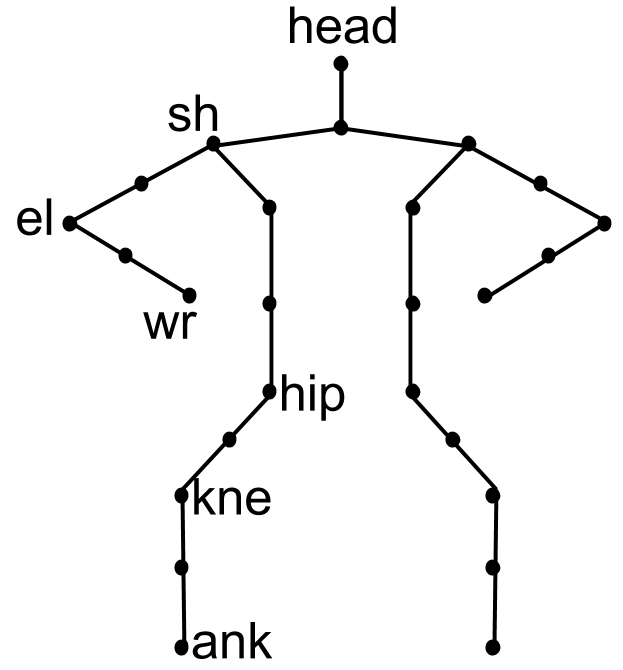}}  \ \ \ \ \
  \subfigure[]{\label{fig:LCT}
  \includegraphics[width=0.45\linewidth]{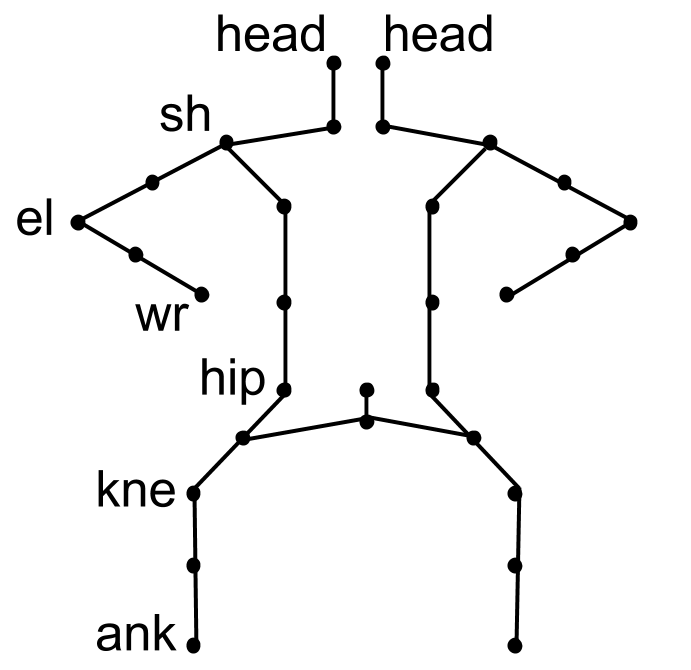}}
\end{center}
   \caption{(a) Tree used by
~\cite{felzenszwalb2005pictorial, yang2011articulated},
the \emph{Upper-Constrained Tree}. (b) the \emph{Lower-Constrained Tree}.}
\label{fig:long}
\label{fig:Trees}
\end{figure}

The space of all possible poses, $(\mathbf{p},\mathbf{t})$, is combinatorially
large. Therefore, to search for a pose that maximizes the score,
$S(I,\mathbf{p},\mathbf{t}; \mathbf{w},\mathbf{b})$, is very hard. However, if
the connections between the parts are
restricted to a tree-structure, the maximization of $S(I,\mathbf{p},\mathbf{t};
\mathbf{w},\mathbf{b})$ over this space can be performed efficiently using
Dynamic
Programming~\cite{felzenszwalb2005pictorial, yang2011articulated}. The structure
of the tree that is generally used is as shown in the
Fig~\ref{fig:UCT} where strongest interactions are taken to
be the kinematic constraints between the adjacent parts in the human body. The
downside of such a restriction is that many other interactions
among the parts remain unhandled, limiting the expressive power of the model.
Two commonly occurring, yet significant manifestations of such unhandled part
interactions
are (a) self-occlusion and (b) a confusion in the localization of
unconnected symmetric parts.

Self-occlusion occurs when one part of the body partially or fully occludes
another
part. Due to heavy articulations in the human body, almost any part can
potentially occlude any other part. This leads to a large variation in the
appearance of a part caused by other parts that may not be directly connected to
it in the tree.
For example, a fully visible upper arm appears very different than an upper
arm partially occluded by the torso~\ref{fig:soFMoP}. Similarly, a fully
visible head appears very different than a head occluded by a lifted arm.

Another important consequence of limiting the part-interactions to a tree
structure is the confusion in the localization of the unconnected symmetric
limbs. For example, in the tree shown in Fig~\ref{fig:UCT}, no kinematic
constraint exists between the left and the right legs. This often leads to an
overlap of their predictions with one of the legs (Fig~\ref{fig:loFMoP}). On the
other hand, since the left and the right arms are constrained through the
shoulder connections, their locations are much more accurately predicted in
this tree structure.

\subsection{Phraselet Clustering}
\begin{figure}[h]
\begin{center}
  \subfigure[]{\label{fig:OP1}\includegraphics[
  width=0.3\linewidth]{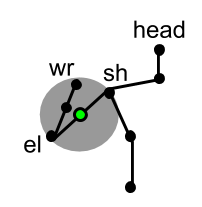}} \
  \subfigure[]{\label{fig:OP2}\includegraphics[
  width=0.3\linewidth]{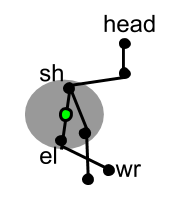}} \
  \subfigure[]{\label{fig:OP3}\includegraphics[
  width=0.3\linewidth]{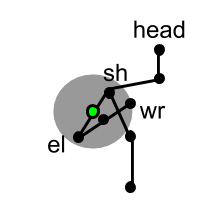}} \
\end{center}
   \caption{Overlap patterns for the upper-arm. sh: shoulder, el: elbow, wr:
wrist.}
\label{fig:ovpat}
\end{figure}

Self-occlusion leads to an overlap of the parts in an image,
thus creating a change in the observed appearance of the parts. Desai and
Ramanan~\cite{desai2012detecting} handle these variations by learning a mixture
of templates for every
part. These templates capture the appearances of different clusters of overlap
patterns, called Phraselets. An overlap pattern for part $i$ is
represented by the relative placements of parts that are in close vicinity of
$i$. For example,
Fig~\ref{fig:ovpat} shows three overlap patterns around the upper-arm.
Fig~\ref{fig:OP1} and Fig~\ref{fig:OP2} are differentiated since the parts close
to the upper-arm are different, while Fig~\ref{fig:OP1} and Fig~\ref{fig:OP3}
are differentiated since the the same parts that are close to the upper-arm are
differently placed around it.

\begin{figure*}[t]
\begin{center}
  \subfigure[Phraselets for the right elbow]{\label{fig:Phr}\includegraphics[
  width=1.0\linewidth]{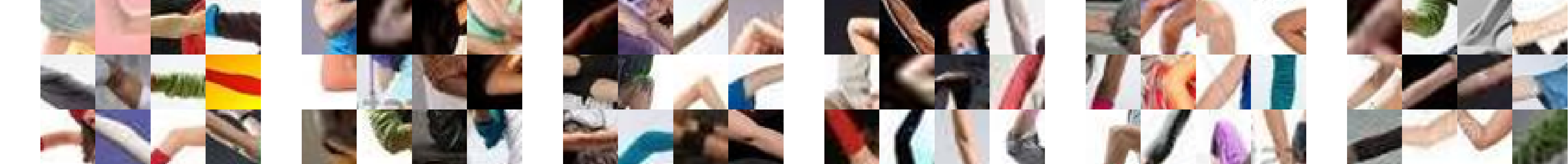}} \\
  \subfigure[Rotation Normalized Phraselets for the right
elbow]{\label{fig:RotPhr}\includegraphics[
  width=1.0\linewidth]{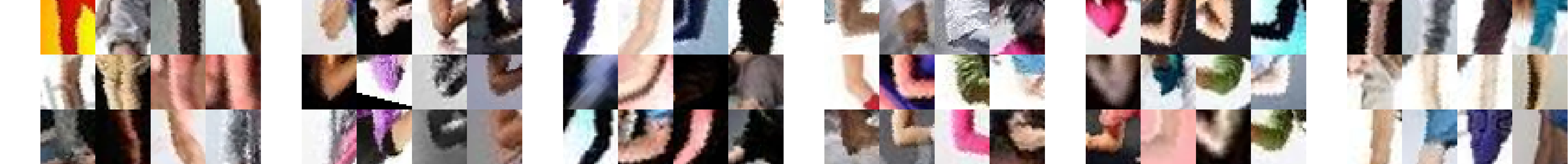}}  
\end{center}
   \caption{Clusters of overlap patterns around the right elbow by Phraselets
and Rotation Normalized Phraselets.}
\label{fig:Clusts}
\end{figure*}

Formally, an overlap pattern for a part $i$ is expressed as a vector of weighted
relative placements of the other parts, where the weighting is based on the
distance of the parts from the part $i$; smaller the distance, larger the
weight and vice versa. Mathematically, it can be written as, $\boldsymbol\Delta
= [\boldsymbol\Delta_1^T \ldots \boldsymbol\Delta_{|V|}^T]^T$, where
$\boldsymbol\Delta_{j} = exp(-\|\boldsymbol\delta_{j}\|)\cdot
(\boldsymbol\delta_{j})$, $\boldsymbol\delta_j = [x_j-x_i, \ y_j-y_i]^T$ and
$|V|$ is the number of parts. For all the training images,
$\boldsymbol\Delta$'s are formed and similar overlap patterns are clustered
together using k-means clustering. These clusters are the Phraselets for
the part $i$. A mixture of k templates are learned for the k Phraselets of the
part $i$. Phraselets for the right elbow are shown in Fig~\ref{fig:Phr}.

In the following sections, we present our improvements on Desai and Ramanan's
Phraselets~\cite{desai2012detecting} in handling self-occlusion, followed by our
proposed solution for the confusion in the localization of non-adjacent
symmetric part.

\section{Improvements in Handling Self-Occlusion}


\begin{figure}[]
\begin{center}
  \subfigure[]{\label{fig:OP1_do}\includegraphics[
  width=0.3\linewidth]{PhraseletClust1.png}} \ \ \ \ \ \ \ \ \
  \subfigure[]{\label{fig:OP2_do}\includegraphics[
  width=0.3\linewidth]{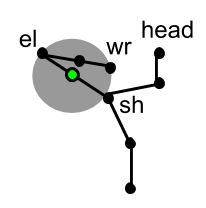}} 
\end{center}
   \caption{Similar overlap patterns at different orientations for the
upper-arm. sh: shoulder, el: elbow, wr: wrist.}
\label{fig:ovpat_do}
\end{figure}

Consider the overlap patterns around the upper-arm in Fig~\ref{fig:OP1_do} and
Fig~\ref{fig:OP2_do}. It can be seen that the patterns are very similar, except
that they are differently oriented. However, due to the rotation, the relative
placements of the parts are different. Therefore, these patterns are placed in
different clusters and different templates are learned for them. 
This method works well if the patterns repeat enough number of times at many
different orientations in the training set such that templates can be learned
for each orientation. 
However, this would require very large training sets which may not always be
available.

We note that the data efficiency of Phraselets can be significantly improved by
representing the overlap patterns in an orientation agnostic manner. Towards
this,
we specify an overlap pattern for part $i$ by the relative placements of parts
that are in close vicinity, normalized according to the
orientation of $i$. More precisely, we modify the weighted relative
placement, $\boldsymbol\Delta_{j}$, to weighted rotation-normalized relative
placement, $\boldsymbol\Delta_{j} = exp(-\|\boldsymbol\delta_{j}\|)\cdot
(R_{-\theta_i} \boldsymbol\delta_{j})$, where $\theta_i$ is the orientation of
part $i$ and $R_{\theta}$ is the rotation matrix for angle $\theta$.

\begin{figure}[]
\begin{center}
   \includegraphics[width=1.0\linewidth]{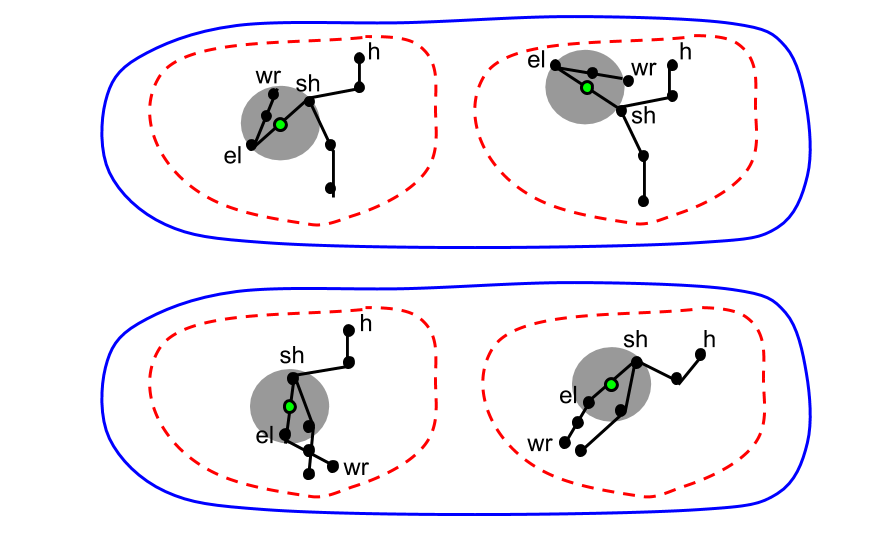}
\end{center}
   \caption{The red dashed lines depict the Phraselet
clustering~\cite{desai2012detecting} for the upper-arm, while the blue solid
lines depict the Rotation Normalized Phraselet clustering.}
\label{fig:long}
\label{fig:RotNorm}
\end{figure}

Further, we note that appearance variation of a part is generally
caused by only a few other parts that often happen to come physically close to
it. These are determined by defining a set of occluding parts, $\mathcal{O}_i$,
as the set of parts that overlap with $i$ in at least $m$ (=100) training
images. For every positive training image,  $\boldsymbol\Delta =
[\boldsymbol\Delta_1^T
\ldots \boldsymbol\Delta_{|\mathcal{O}_i|}^T]^T$ is formed, where
$|\mathcal{O}_i|$ is the number of occluding parts of part $i$. As before, the
set of $\boldsymbol\Delta$'s obtained from the training images is clustered
using k-means and a template is trained per cluster. Now, each cluster
represents a Rotation Normalized Phraselet. Occluding parts facilitate
formation of cleaner clusters. Fig~\ref{fig:RotNorm} schematically
shows the difference in clustering for the upper-arm between
Phraselets~\cite{desai2012detecting} and the Rotation
Normalized Phraselets. Fig~\ref{fig:RotPhr}  shows our Rotation Normalized
Phraselets of the right elbow compared with the Phraselets in Fig~\ref{fig:Phr}.
It can be observed that, due to rotation normalization, a wider variety of
overlap patterns are captured with much less repetition of patterns across the
clusters.

Since the templates are now rotation normalized, they are matched at all
orientations. Therefore, $p_i$ is updated to: $p_i=(x_i,y_i,\theta_i)$, where
$\theta_i$ is the orientation of part $i$. Also,
$\psi(p_i-p_j)=[-dx \ -dx^2 \ -dy \ -dy^2 \ cos(d\theta)]$, with $dx
=(x_i-x_j-\mu
x_{i,j}^{t_i,\theta_i})$. $\mu x_{i,j}^{t_i,\theta_i}$ is calculated as $\mu
x_{i,j}^{t_i,\theta_i} = \mu r_{i,j}^{t_i} cos(\theta_i)$, where $\mu
r_{i,j}^{t_i}$ is the average distance between the part $i$ and its parent $j$
in the training images with type $t_i$ for part $i$. During inference,
$(\mathbf{p}^*,\mathbf{t}^*) = argmax \
S(I,\mathbf{p},\mathbf{t};\mathbf{w},\mathbf{b})$ is calculated. Learning is
performed using (\ref{eqn:SSVM}).

\section{Localization of Non-adjacent Symmetric Parts}
Often, predicted locations of the symmetric parts overlap with one
of the parts if they are not kinematically constrained in the part-based model.
For example, in Fig~\ref{fig:loFMoP}, the leg predictions from the tree of
Fig~\ref{fig:UCT} overlap with the same leg due to the absence of a constraint
between them. However, enforcing the constraints between both the shoulders as
well as the hips would introduce a cycle in the tree, inhibiting efficient
maximization using Dynamic Programming~\cite{felzenszwalb2005pictorial}. 

Towards unambiguous localization of the symmetric lower limbs while still
exploiting the
advantage of tree structured interactions, we define a new tree as shown in
Fig~\ref{fig:LCT}. In this tree, the connection
between the left and right side happens in the lower body and not in the upper
body. Specifically, two nodes are added around the pelvis region. This
connection constrains the left and the right lower limbs kinematically and
avoids the overlap of leg predictions. Furthermore, it strengthens the
localization of the lower body parts by collecting additional image evidence
around the pelvis region. We refer to this tree as the \emph{Lower-constrained
Tree} and the traditionally used tree (shown in Fig~\ref{fig:UCT}) as the
\emph{Upper-constrained Tree}.

A missing constraint in the upper part of the \emph{Lower-constrained Tree} may
produce erroneous upper body pose estimates, consequently affecting the lower
body pose estimation as well. We observe that the head has a distinctive
appearance
and is localized reliably. Therefore, in order to mimic the constraints in the
upper body, we introduce two head nodes in the \emph{Lower-constrained Tree} as
shown in Fig~\ref{fig:LCT}. This is especially effective when there is less
ambiguity in the head location. In such cases, the \emph{Lower-constrained Tree}
outperforms the \emph{Upper-constrained Tree}, even for the upper parts; the
correct localization of the lower parts along with unambiguous head location
helps in the localization of the upper parts as well. However, in general, the
\emph{Lower-constrained Tree} still underperforms on the upper parts. Due
to this, we combine the two complementing trees in order to obtain superior pose
estimates for both the upper and the lower parts.

Our method of combining the two trees is based on the following idea: suppose we
know the true locations of the lower body parts in an image. Now, if we infer
the upper body pose using the \emph{Upper-constrained Tree} with the lower
parts fixed to their true locations, the estimates of the upper body parts would
improve. However, in practice, we do not know the true locations. Instead,
if we use a different model that performs better lower body pose estimation,
the overall accuracy increases.

\begin{figure}[t]
 \begin{center}
  \includegraphics[width=0.9\linewidth]{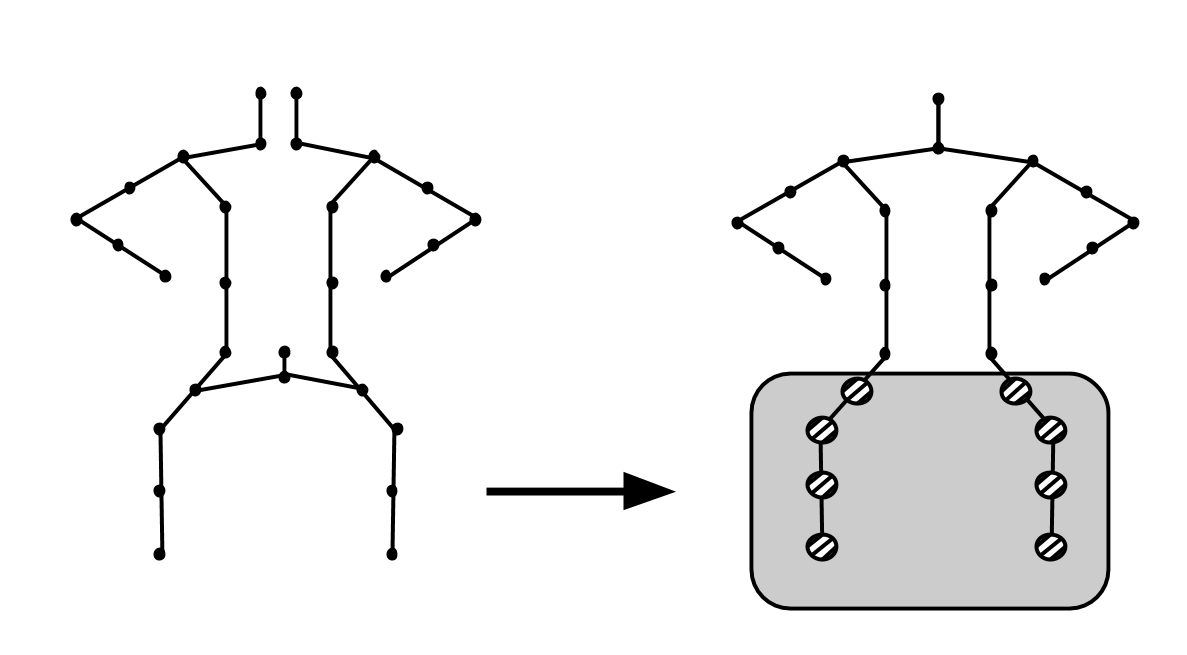}
 \end{center}
   \caption{Two Trees pose estimation.}
\label{fig:Pl}
\end{figure}

\textbf{Two Trees Pose Estimation} (Fig~\ref{fig:Pl}):
First, the pose is estimated using the \emph{Lower-constrained Tree}. Treating
the predicted pose of just the lower-body parts as the input evidence, the upper
body pose is again estimated using the \emph{Upper-constrained Tree}.

Similar combination can be performed by embedding the upper body pose predicted
using the \emph{Upper-constrained Tree} into the \emph{Lower-constrained Tree}.
However, due to the constraint in the lower part and the two heads nodes
mimicking the constrains in the upper part, the \emph{Lower-constrained Tree}
generally localizes the parts more accurately. Therefore, starting with the
\emph{Upper-constrained Tree} and re-estimating with the \emph{Lower-constrained
Tree} does not generally perform better than the opposite way.
The \emph{Upper-constrained Tree} and the \emph{Lower-constrained Tree} are
trained independently using (\ref{eqn:SSVM}). 

Note that Wang and Mori~\cite{wang2008multiple} also use multiple complementing
trees to handle the non-adjacent part interactions. However, the difference is
mainly in combining the trees. While they combine the distribution over
poses returned by the trees using boosting, we combine the max-scoring
poses returned by the trees using our Two Trees Pose Estimation.

\section{Experimental Evaluation}
\subsection{Setup}
\subsubsection{Datasets} 
We use three datasets, namely the Leeds Sports (LSP) dataset~\cite{Johnson10},
the Image Parse (IP) dataset~\cite{ramanan2006learning} and a Dance dataset
created by us. All these datasets have challenging poses with complex part
interactions. Hence they are suitable to evaluate our ideas.

The LSP has 1000 images for training and
1000 images for testing. It contains images of people involved in sports
like football, gymnastics, tennis etc and is particularly challenging in terms
of pose variations. The annotations contain pixel locations of 14 keypoints
(head, neck, shoulders, elbows, wrists, hips, ankles and knees). The IP has 100
images for training and 205 images for testing. The nature of images and the
annotated keypoints are similar to that of LSP.

The Dance dataset has 150 images for training and 72 images for testing. The
images are selected from the top searches returned by Google for the keywords
''Dance Solo``, ''Hip hop poses`` and ''Modern dance poses``. Similar to LSP,
14 keypoints are annotated for each image.

\subsubsection{Evaluation Measure}
We use the standard Percentage of Detected Joints
(PDJ)~\cite{sapp2013modec, toshev2013deeppose} as our
evaluation measure. According to PDJ, a joint is correctly
detected if the distance between the predicted joint location and the
groundtruth location is within some fraction of the torso diameter (distance
between the left shoulder and the right hip). In a PDJ curve, this fraction is
varied between 0 and 0.5 and the percentage of detected joints are plotted for
each value of this fraction~\cite{toshev2013deeppose}. $PDJ_{avg}$ is used to
represent the average percentage of detected joints over the whole curve.

\subsubsection{Implementation Details}
For all our experiments, we use 7 part-types and 36 orientations. For
fairness in comparison, we retrain the MoP~\cite{yang2011articulated} and the
Phraselets~\cite{desai2012detecting} with 7 part-types as well. With most part
of the code in matlab, our full model (Rotation Normalized Phraselets + Two
Trees Pose Estimation) takes about 15 seconds using 8 parallel threads and about
60 seconds using a single thread on a typical image in the LSP dataset.

\subsection{Results and Discussion}
In this section, we first analyze the two orthogonal solutions that we propose.
Then, we combine both the orthogonal solutions and compare our results with the
state-of-the art.

\begin{figure}
\begin{center}
    \subfigure[Pose
estimates of Phraselet~\cite{desai2012detecting}]{\label{fig:Phr_qual}
  \includegraphics[width=1.0\linewidth]{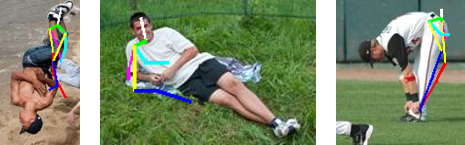}} \\

    \subfigure[Pose estimates of Rotation Normalized
Phraselet]{\label{fig:RotNormPhr_qual}
  \includegraphics[width=1.0\linewidth]{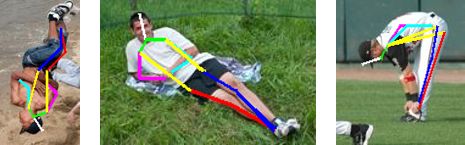}}
  \end{center}
   \caption{Qualitative analysis of Rotation Normalized Phraselet.}
\label{fig:RotNorm_qual}
\end{figure}

\begin{table}
\begin{center}
\begin{tabular}{|c|c|r|r|r|r|}
\hline
Category &  & \multicolumn{1}{c|}{Elb} & \multicolumn{1}{c|}{Wri} &
\multicolumn{1}{c|}{Kne} & \multicolumn{1}{c|}{Ank} \\ \hline
\multicolumn{ 1}{|c|}{Sports} & MoP~\cite{yang2011articulated} & 38.4 & 30.8 &
46.2 & 43.2 \\ \cline{ 2- 6}
\multicolumn{ 1}{|c|}{} & Phraselets~\cite{desai2012detecting} & 40.2 & 30.6 &
47.8 & 45.2 \\ \cline{ 2- 6}
\multicolumn{ 1}{|c|}{} & \shortstack{RotNorm \\ Phr (Ours)} &
\textbf{48.3} & \textbf{37.5} & \textbf{51.8} & \textbf{47} \\ \hline
\multicolumn{ 1}{|c|}{Gym} & MoP~\cite{yang2011articulated} & 15.2 & 12.1 & 14.6
& 15.8 \\ \cline{ 2- 6}
\multicolumn{ 1}{|c|}{} & Phraselets~\cite{desai2012detecting} & 16 & 10.2 &
14.9 & 16.4 \\ \cline{ 2- 6}
\multicolumn{ 1}{|c|}{} & \shortstack{RotNorm \\ Phr (Ours)} &
\textbf{22.3} & \textbf{18.4} & \textbf{21.6} & \textbf{18.4} \\ \hline
\multicolumn{ 1}{|c|}{Dance} & MoP~\cite{yang2011articulated} & 52.9 & 43.9 &
48.3 & 39.7 \\ \cline{ 2- 6}
\multicolumn{ 1}{|c|}{} & Phraselets~\cite{desai2012detecting} & 52.4 & 44.8 &
45.6 & 39.1 \\ \cline{ 2- 6}
\multicolumn{ 1}{|c|}{} & \shortstack{RotNorm \\ Phr (Ours)} &
\textbf{67.4} & \textbf{61.8} & \textbf{57.8} & \textbf{49.6} \\ \hline
\end{tabular}
\end{center}
\caption{$PDJ_{avg}$ values for various categories of images.}
\label{tab:RotNorm}
\end{table}

\subsubsection{Rotation Normalized Phraselets}
Recall that Rotation Normalized Phraselets capture the appearances of overlap
patterns around a part normalized according to the part's orientation while
Phraselets~\cite{desai2012detecting} capture the appearances of the unnormalized
overlap patterns.
Due to limited training data, the overlap patterns generally do not
repeat enough number of times at multiple orientations such that a template can
be learned for each orientation. Fig~\ref{fig:RotNorm_qual} shows
some cases where Phraselets fail to handle the non-upright self-occlusion
patterns while the Rotation Normalized Phraselets are able to estimate the pose
correctly.

For quantitative evaluation of Rotation Normalized Phraselets, we form three
categories of images, namely Sports, Gym and Dance. The Sports category contains
all the images from the LSP test set. The Gym category is formed by manually
selecting the images from the LSP test set if the activity in the image is
gymnastics or aerobics. There are 129 images in the Gym category. All the images
in the Dance test set are placed in the Dance category.

We evaluate three algorithms, namely the basic Mixture of
Parts~\cite{yang2011articulated} (MoP) model,
Phraselets~\cite{desai2012detecting} and our Rotation Normalized Phraselets on
all the image categories. For evaluation on Sports and Gym categories, models
are trained on the LSP training set, while for evaluation on Dance category,
models are trained on the Dance training set. The $PDJ_{avg}$ values of elbow,
wrist, knee and ankle are reported in Table~\ref{tab:RotNorm}. It can be seen
that Phraselets
modestly improves over MoP while Rotation Normalized Phraselets obtain large
improvement on all the parts. The gain in performance is especially significant
on more challenging categories such as Gym and Dance.

\begin{figure}[h]
\begin{center}
    \subfigure[Pose estimates from the \emph{Upper-constrained
Tree}]{\label{fig:UCT_qual}
  \includegraphics[width=0.7\linewidth]{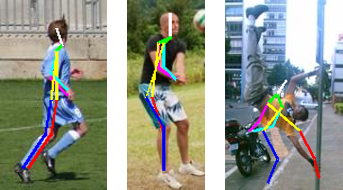}} \\
    \subfigure[Pose estimates from the
\emph{Lower-constrained Tree}]{\label{fig:LCT_qual}
  \includegraphics[width=0.7\linewidth]{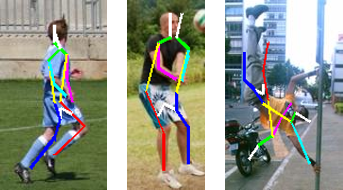}} \\
    \subfigure[Two Trees Pose Estimation]{\label{fig:2T_qual}
  \includegraphics[width=0.7\linewidth]{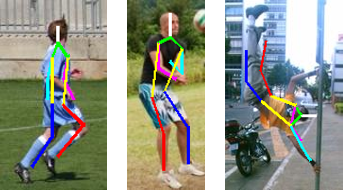}}
  \end{center}
   \caption{Qualitative analysis of Rotation Normalized Phraselet.}
\label{fig:long}
\label{fig:2Trees_qual}
\end{figure}

\begin{table}
\begin{center}
\begin{tabular}{|c|c|r|r|r|r|r|}
\hline
\textbf{Meth.} &  & \multicolumn{1}{c|}{Sho} & \multicolumn{1}{c|}{Elb} &
\multicolumn{1}{c|}{Wri} & \multicolumn{1}{c|}{Kne} & \multicolumn{1}{c|}{Ank}
\\ \hline
\multicolumn{ 1}{|c|}{MoP~\cite{yang2011articulated}} & Up T & 51.8 & 38.4 &
30.7 & 46.2 & 43.1 \\ \cline{ 2- 7}
\multicolumn{ 1}{|c|}{} & Lo T & \textbf{54} & \textbf{40.3} & \textbf{32} &
49.7 & 46.5 \\ \cline{ 2- 7}
\multicolumn{ 1}{|c|}{} & 2T & 53.3 & 39.3 & 31.5 & \textbf{49.7} &
\textbf{46.5} \\ \hline
\multicolumn{ 1}{|c|}{Phr~\cite{desai2012detecting}} & Up T & 56.2 & 40.2 &
30.6 & 47.8 & 45.2 \\ \cline{ 2- 7}
\multicolumn{ 1}{|c|}{} & Lo T & 58 & \textbf{42.2} & \textbf{33.5} & 52.1 &
48.7 \\ \cline{ 2- 7}
\multicolumn{ 1}{|c|}{} & 2T & \textbf{58.3} & 41.7 & 31.7 & \textbf{52.1} &
\textbf{48.7} \\ \hline
\multicolumn{ 1}{|c|}{RotNorm} & Up T & \textbf{60.3} & \textbf{48.3} &
37.5 & 51.8 & 47 \\ \cline{ 2- 7}
\multicolumn{ 1}{|c|}{Phr (Ours)} & Lo T & 55 & 44.9 & 36 & 55.4 & 50 \\ \cline{
2- 7}
\multicolumn{ 1}{|c|}{} & 2T & 60.1 & 48.1 & \textbf{37.9} &
\textbf{55.4} & \textbf{50} \\ \hline
\end{tabular}
\end{center}
\caption{Two Trees Pose Estimation Analysis on the LSP dataset~\cite{Johnson10}.
Up T: \emph{Upper-constrained Tree}, Lo T: \emph{Lower-constrained Tree}, 2T:
Two Trees Pose Estimation.}
\label{tab:TreeModels}
\end{table}

\subsubsection{Two trees Pose Estimation} 
Fig~\ref{fig:2Trees_qual} shows some pose estimates from the
\emph{Upper-constrained Tree}, the \emph{Lower-constrained Tree} and the
combination of the two trees. In
the first and the second images, head appearance is distinctive. Therefore,
the two head nodes in the \emph{Lower-constrained Tree} (Fig~\ref{fig:LCT})
localize accurately, effectively mimicking the constrains in the upper body.
This, in addition to the
constraints in the lower body causes the \emph{Lower-constrained Tree} to
localize both the upper and lower parts correctly by itself. The second image
also shows how an erroneous lower body pose estimation in the
\emph{Upper-constrained Tree} (Fig~\ref{fig:UCT}) can cause an error in the
upper body pose estimation as well. The third image shows an example where the
\emph{Upper-constrained Tree} predicts an inaccurate pose, the
\emph{Lower-constrained Tree} predicts a partially accurate pose but the
combination predicts a fully accurate pose. 

Table~\ref{tab:TreeModels} reports the $PDJ_{avg}$ values of various joints
for the \emph{Upper-constrained Tree}, the \emph{Lower-constrained Tree} and the
two trees
for the three algorithms on the LSP dataset. First, we point out that the
\emph{Lower-Constrained Tree} consistently improves the localization
accuracy of the lower parts. Further, the Two Trees Pose Estimation shows
superior localization accuracies for both the upper and the lower body parts.
Note that this is achieved by only doubling the running time. 

An interesting observation is that in algorithms where there is no rotation
normalization, such as MoP~\cite{yang2011articulated} and
Phraselets~\cite{desai2012detecting},
the \emph{Lower-Constrained Tree} performs better on all the parts in
comparison with the \emph{Upper-Constrained Tree}. This is because, when there
is no rotation normalization, the templates are searched only over translations
and scales. Due to the reduced search space, the ambiguity in head location
reduces. Therefore, both the head nodes in the \emph{Lower-constrained Tree}
localize accurately just based on the appearance, leading to a
better localization of the upper body parts as well. This observation is
very useful for the algorithms without rotation normalization, since pose
estimation on the \emph{Lower-constrained Tree} takes
the same time as the traditionally used tree (the \emph{Upper-Constrained
Tree}), yet being more accurate.

\begin{figure}[h]
\begin{center}
  \subfigure[elbow]{\label{fig:avg_image_inr}\includegraphics[
  width=0.48\linewidth]{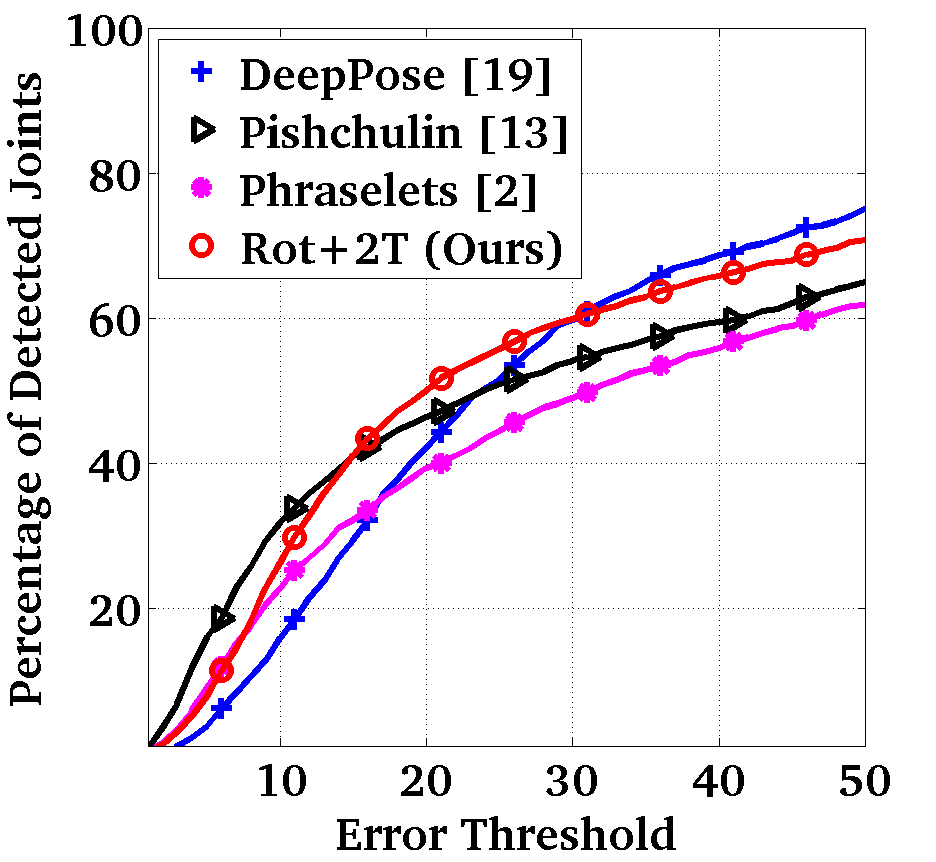}}
    \subfigure[wrist]{\label{fig:avg_image_daim}
  \includegraphics[width=0.48\linewidth]{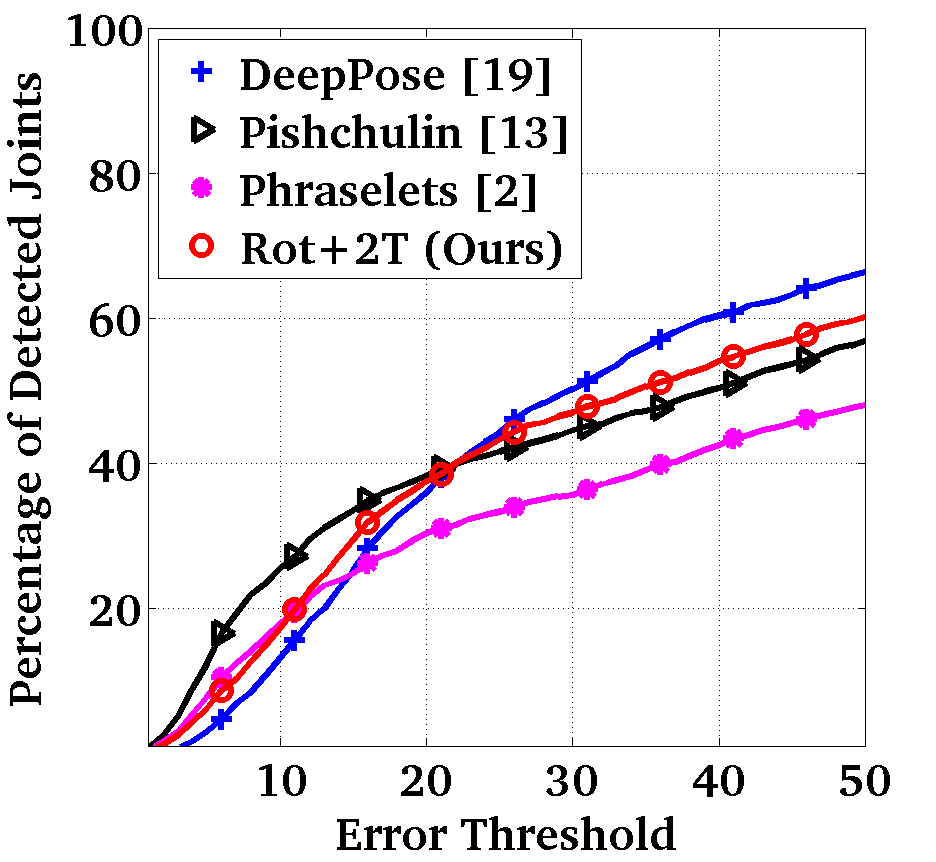}} \\

  \subfigure[knee]{\label{fig:avg_image_inr}\includegraphics[
  width=0.48\linewidth]{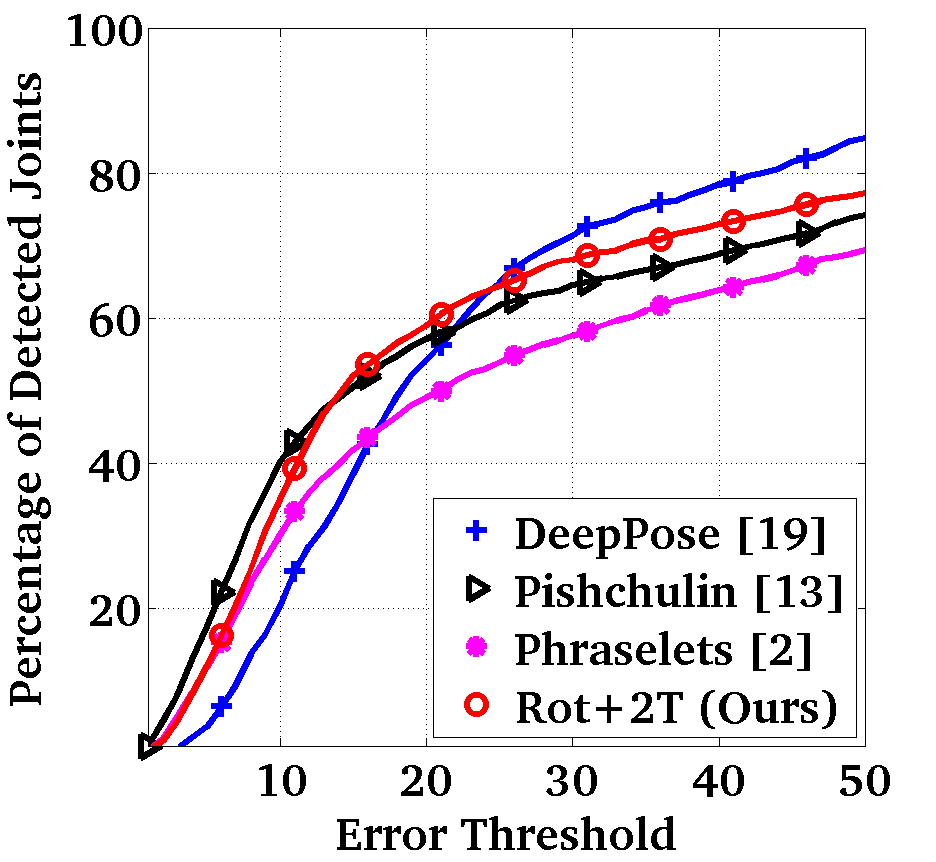}}
    \subfigure[ankle]{\label{fig:avg_image_daim}
  \includegraphics[width=0.48\linewidth]{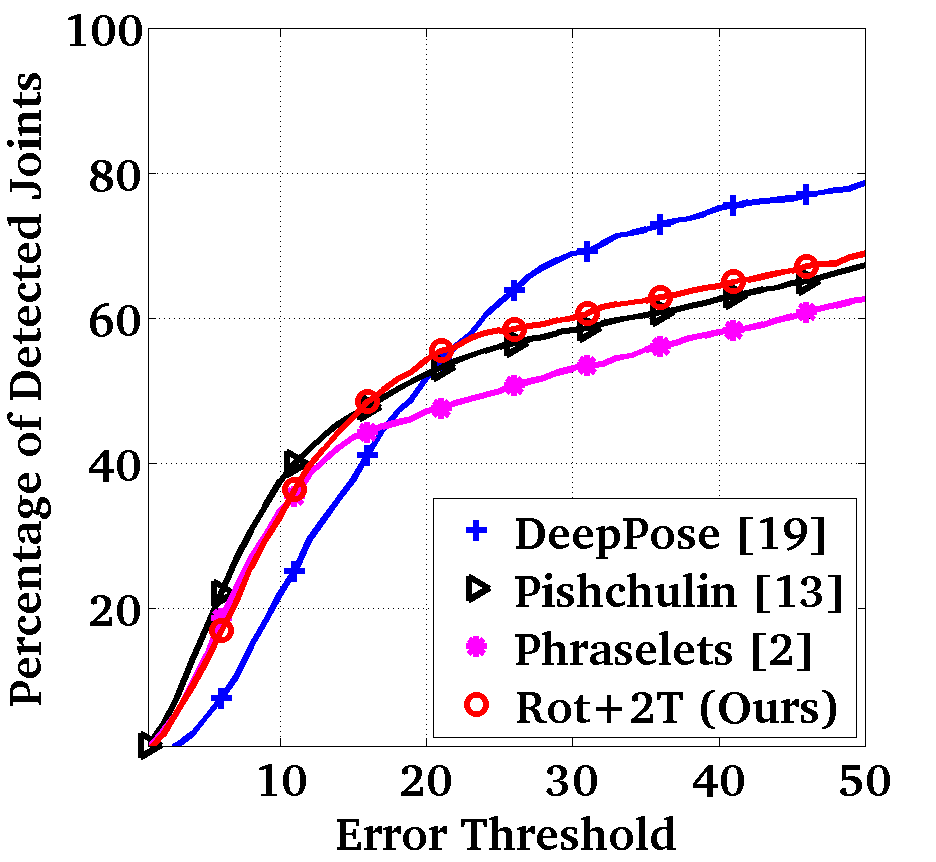}}
  \end{center}
   \caption{PDJ curves on the LSP dataset~\cite{Johnson10} (best viewed in
color).}
\label{fig:long}
\label{fig:comparisonLSP}
\end{figure}

\begin{figure}[h]
\begin{center}
  \subfigure[elbow and wrist]{\label{fig:avg_image_inr}\includegraphics[
  width=0.48\linewidth]{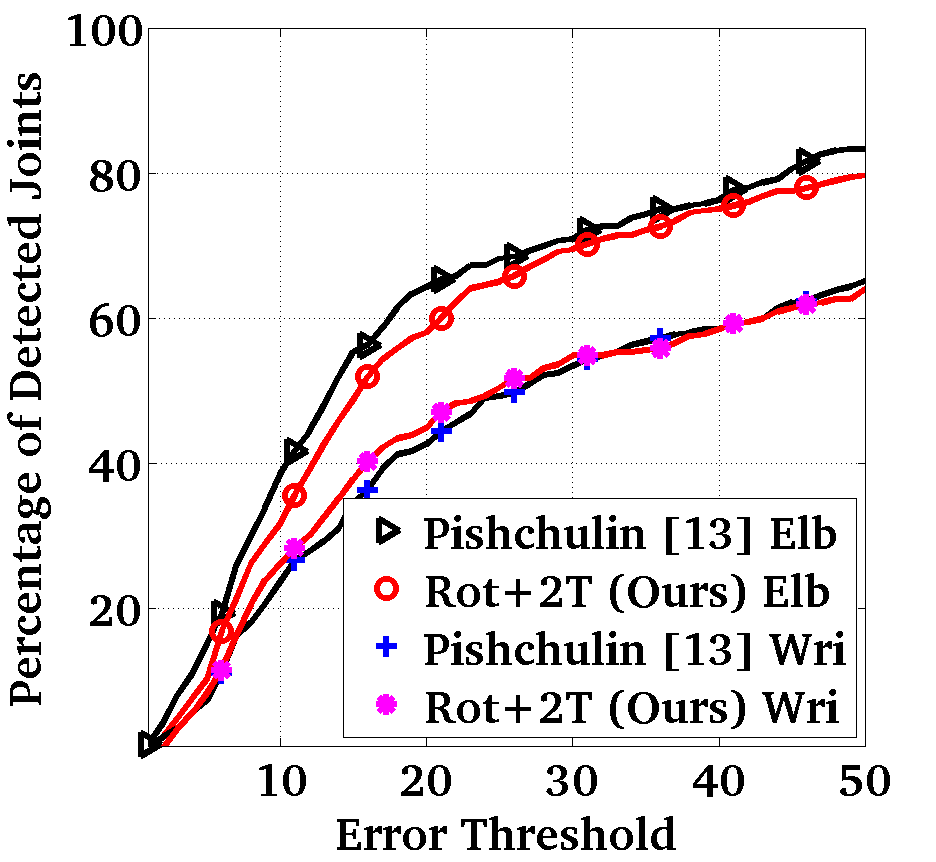}}
    \subfigure[Knee and ankle]{\label{fig:avg_image_daim}
  \includegraphics[width=0.48\linewidth]{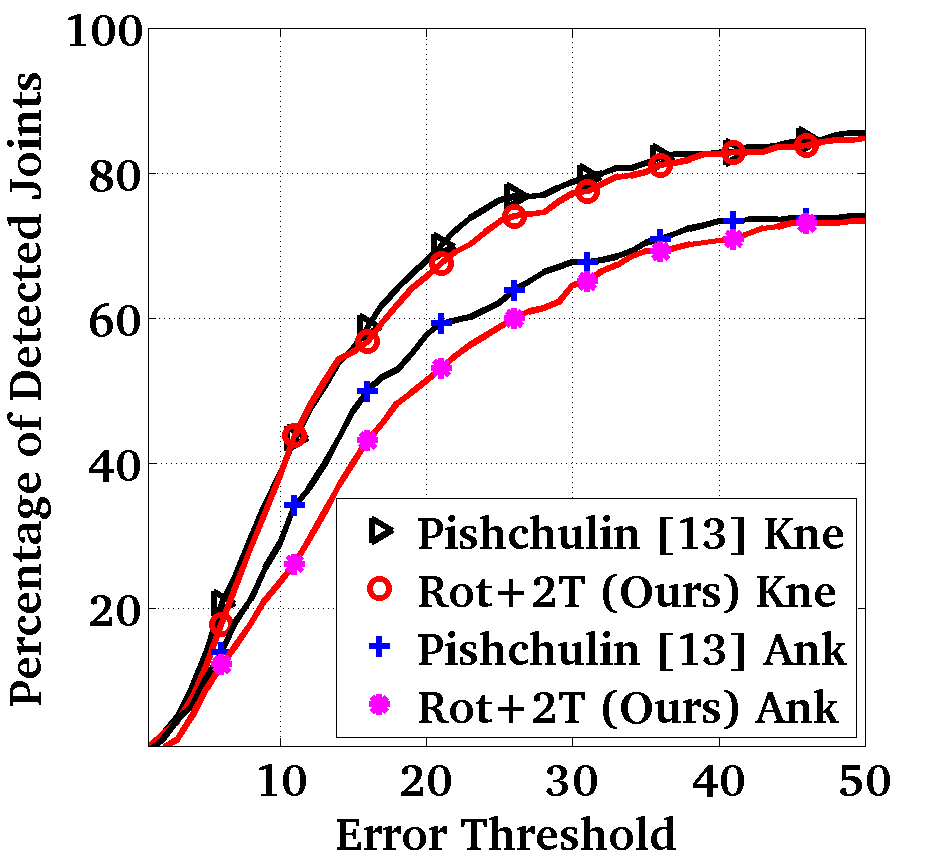}}

  \end{center}
   \caption{PDJ curves on the Image Parse dataset~\cite{ramanan2006learning}
(best viewed in color).}
\label{fig:long}
\label{fig:comparisonIP}
\end{figure}

\subsubsection{Comparisons with the Other Methods}
We note that our best performing model is a combination of the Rotation
Normalized Phraselets and the Two Trees Pose Estimation reported in the last
row of Table~\ref{tab:TreeModels}. We compare this model with the
state-of-the-art methods on two datasets, namely the LSP~\cite{Johnson10} and
the IP~\cite{ramanan2006learning}.
Comparative PDJ curves on the LSP dataset are shown in
Fig~\ref{fig:comparisonLSP}. It can be  seen
that we beat the baseline (Desai and Ramanan~\cite{desai2012detecting}) by a
large margin. Also, no method out-performs all the methods and we perform
equivalently to the best performing methods.

Our performance is very similarly to that of the state-of-the-art part-based
model of Pishchulin \etal~\cite{pishchulin2013strong}. They combine the
best practices in the Human Pose Estimation problem and show that such a
combination can produce very good results. However, their method uses many
complex DPM and shape context templates at various granularities to perform
accurate localization. Due to this, the running time of their released
code on a typical image in the
LSP test set increases to about 7 minutes. On the other hand, we use
simple part-level HOG templates and achieve a similar performance with a running
time of about 15 seconds on the same images. Moreover, the
specialized detectors of Pishchulin \etal~\cite{pishchulin2013strong} can be
used with our model as well to boost our performance, albeit at the cost of
increased running time.

It can be observed that DeepPose~\cite{toshev2013deeppose} gives good results
when the error threshold for detection of joints is high (right half of the PDJ
curves), while it underperforms when the threshold is low. This implies that,
in comparison with the other methods, DeepPose makes many approximate
predictions, but fails to localize the parts accurately.
This is expected from a pure learning based approach that does not model
the articulations explicitly.

We also compare our method on the IP dataset. The PDJ curves are shown in
Fig~\ref{fig:comparisonIP}. It can be seen that our performance is similar to
that of Pishchulin \etal~\cite{pishchulin2013strong}, albeit again at a
much lower computational cost.

\section{Conclusion}
Tree structured pair-wise constraints are restrictive in terms of encoding all
the possible part interactions. Two strong manifestations of such unhandled
part interactions are self-occlusion among the parts and
a confusion in the localization of the non-adjacent symmetric limbs. We
propose two modular and efficient improvements to Desai and
Ramanan's~\cite{desai2012detecting} method to address the above problems.
First, we propose Rotation Normalized Phraselets for handling
self-occlusion in a more data efficient manner. We show especially large
improvements on uncommon poses such as sports, gymnastics and dance. Secondly,
we propose a solution for handling the confusion in the localization of
non-adjacent symmetric limbs using a combination of two complementing trees and
report a boost in performance taking only twice the time. We also show that a
combination of the above two solutions improves the results compared to either
used alone. We evaluate our method on two standard datasets and achieve
equivalent results to the best performing methods in much less time when
compared to the state-of-the-art part based model.

{\small
\bibliographystyle{ieee}
\bibliography{egbib}
}

\end{document}